\documentclass[conference]{IEEEtran}
\pdfoutput=1
\IEEEoverridecommandlockouts
\usepackage{cite}
\usepackage{amsmath,amssymb,amsfonts}
\usepackage{algorithmic}
\usepackage{graphicx}
\usepackage{textcomp}
\usepackage{xcolor}

\usepackage{tikz}
\usepackage{subfigure}
\usepackage[ruled,vlined]{algorithm2e}
\usepackage{url}
\usepackage{pstool}

\DeclareMathOperator*{\argmin}{arg\,min}

\newtheorem{remark}{Remark}
\usepackage{bbm}

\def\BibTeX{{\rm B\kern-.05em{\sc i\kern-.025em b}\kern-.08em
    T\kern-.1667em\lower.7ex\hbox{E}\kern-.125emX}}
    
\begin{document}

\newcommand{\norm}[1]{\left\lVert#1\right\rVert}
\newcommand{\todo}[1]{{\color{blue} [TODO: #1]}}

\title{Regularized Classification-Aware Quantization\\
\thanks{This work was funded by Huawei}
}

\author{\IEEEauthorblockN{Daniel Severo}
\IEEEauthorblockA{\textit{Electrical and Computer Engineering} \\
\textit{University of Toronto}\\
Toronto, Canada \\
d.severo@mail.utoronto.ca}
\and
\IEEEauthorblockN{Elad Domanovitz}
\IEEEauthorblockA{\textit{Electrical and Computer Engineering} \\
\textit{University of Toronto}\\
Toronto, Canada \\
elad.domanovitz@utoronto.ca}
\and
\IEEEauthorblockN{Ashish Khisti}
\IEEEauthorblockA{\textit{Electrical and Computer Engineering} \\
\textit{University of Toronto}\\
Toronto, Canada \\
akhisti@ece.utoronto.ca}
}

\maketitle

\begin{abstract}
Traditionally, quantization is designed to minimize the reconstruction error of a data source. When considering downstream classification tasks, other measures of distortion can be of interest; such as the 0-1 classification loss. Furthermore, it is desirable that the performance of these quantizers not deteriorate once they are deployed into production, as relearning the scheme online is not always possible. In this work, we present a class of algorithms that learn distributed quantization schemes for binary classification tasks. Our method performs well on unseen data, and is faster than previous methods proportional to a quadratic term of the dataset size. It works by regularizing the 0-1 loss with the reconstruction error. We present experiments on synthetic mixture and bivariate Gaussian data and compare training, testing, and generalization errors with a family of benchmark quantization schemes from the literature. Our method is called \emph{Regularized Classification-Aware Quantization}.

\end{abstract}

\begin{IEEEkeywords}
distributed quantization; generalization; regularization; classification
\end{IEEEkeywords}

\section{Introduction}
As the worldwide estimate of active edge devices surpasses tens of billions, reducing data storage and transmission costs becomes vital. When dealing with continuous data, most pipelines include quantization, where the data is discretized in some way. Traditionally, quantization is designed to minimize the reconstruction error with respect to some notion of distortion; such as the Mean Squared Error (MSE).

Despite small reconstruction error being an intuitive goal, it might not be the optimal measure when considering downstream tasks such as binary classification \cite{Hanna2019-mi}. For example, there is no need to store data with high precision if it will be thresholded afterwards for decision making. Quantizing near the threshold value will suffice.

Following \cite{Hanna2019-mi}, we assume the fusion center has access to a pre-trained classifier and it wishes to apply it over data features collected at $d$ distributed sensor nodes. The focus of our work is on designing distributed quantizers to be used for classification. Thus, our goal is not to improve the performance of the classifier, but rather, to find a good quantization scheme that would result in minimal performance degradation when used in conjunction with the pre-trained classifier.

The assumption of a pre-trained classifier can be motivated by considering a two phase approach in systems design. When training the classifier, data is sent at high resolution over rate constrained channels, i.e. several channel-uses can be used to send a single sample. After, real-time decisions are required and the data has to be quantized to meet these constraints.

In practical settings, quantization schemes must be learned from a finite set of data points, as the true data generating distribution is not known. In production systems, re-learning with incoming data can be cumbersome, or even impossible, as this would require bidirectional communication between the decoder and encoders to agree on a new code-book. It is therefore desirable for the performance of a learned quantization scheme to not deteriorate once applied to new, but similar, data. This begets the question: \emph{how well do learned quantization schemes generalize to out-of-sample data?}

In this work we propose a quantization scheme called \emph{Regularized Classification-Aware Quantization} (RCAQ) that is learned from data. Experiments show that RCAQ generalizes with few training examples on synthetic Gaussian data. We evaluate RCAQ, and previous work by \cite{Hanna2019-mi} called on-the-line, on both in-sample (training) and out-of-sample (testing) datasets to estimate generalization. Results indicate that both methods have comparable performance. However, RCAQ has significantly lower computational complexity, as it does not require iterating twice over the dataset as in on-the-line. All code is available at \url{https://github.com/dsevero/rcaq}.

\section{Problem Formulation}
This work focuses on a distributed setting, where $d$ encoders must quantize and transmit data to a single decoder tasked with performing linear binary classification. Learning is done with complete information, i.e. the decoder and encoders can communicate to decide on a code-book. Once deployed, encoders do not coordinate with each other and communication with the decoder is unidirectional. Data points are in $\mathbb{R}^d$, but each encoder has access only to one dimension in $\mathbb{R}$. Quantization indices are transmitted losslessly over a noiseless channel. The classifier is considered to be given, i.e. pre-trained.

Our goal is \emph{not} to improve the classifier, but to learn a quantization scheme at the encoders that will minimize the additional classification error introduced by quantization. The quantization error is measured by comparing the output of the classifier with and without quantization. Note that, in some sense, the classifier is our ground truth. 

We extend the formulation of \cite{Rebollo-Monedero2005-uc}, by emphasizing generalization performance. Let $\mathbf{x} = (x_1, \dots, x_d) \in \mathbb{R}^d$ represent a $d$-dimensional datapoint and $y(\mathbf{x}) \in \{-1, +1\}$ the label assigned by the pre-trained classifier \emph{without} quantization. The $k$-th encoder must learn a quantization function $\mathcal{E}_k: \mathbb{R} \rightarrow [2^{R_k}]$, where $R_k$ is the rate and $[2^{R_k}] = \{1, 2, \dots, 2^{R_k}\}$ are the set of integer quantization indices. The decoder receives an integer $d$-tuple $\mathcal{E}(\mathbf{x}) = (\mathcal{E}_1(x_1), \dots, \mathcal{E}_d(x_d))$ and must output a classification $\hat{y}(\mathcal{E}(\mathbf{x})) \in \{-1, +1\}$. 

\begin{remark}
Note that, unlike traditional quantization, there is no need for reconstruction points as the decoder can directly map the quantization integers to a class label.
\label{rem:no-recon}
\end{remark}

If $\mathbf{x} \sim P$ is the true data generating distribution, the objective is to solve the following optimization problem:
\begin{equation}\label{opt-problem}
    \begin{aligned}
        \underset{\mathcal{E}, \hat{y}}{\text{minimize}} & \quad \mathbb{E} [\mathbbm{1}[\hat{y}(\mathcal{E}(\mathbf{x})) \neq y(\mathbf{x})]]\\
        \text{subject to}                                & \quad R_1, \dots, R_d \text{ and } P,
    \end{aligned}
\end{equation}
where the expectation is taken over $\mathbf{x} \sim P$, and $\mathbbm{1}$ the indicator function.

The objective function in equation (\ref{opt-problem}) is known as the \emph{0-1 loss}, and differs from \cite{Hanna2019-mi} only in the choice of $P$. Here, $P$ is the (unknown) data generating distribution, while \cite{Hanna2019-mi} uses the empirical distribution of a set of points $\mathcal{T} = \{\mathbf{x}^{(i)}\}_{i=1}^N$ sampled i.i.d.\ from $P$ (i.e. they optimize the training loss).

\begin{figure}
    \centering
    \includegraphics[width=\columnwidth]{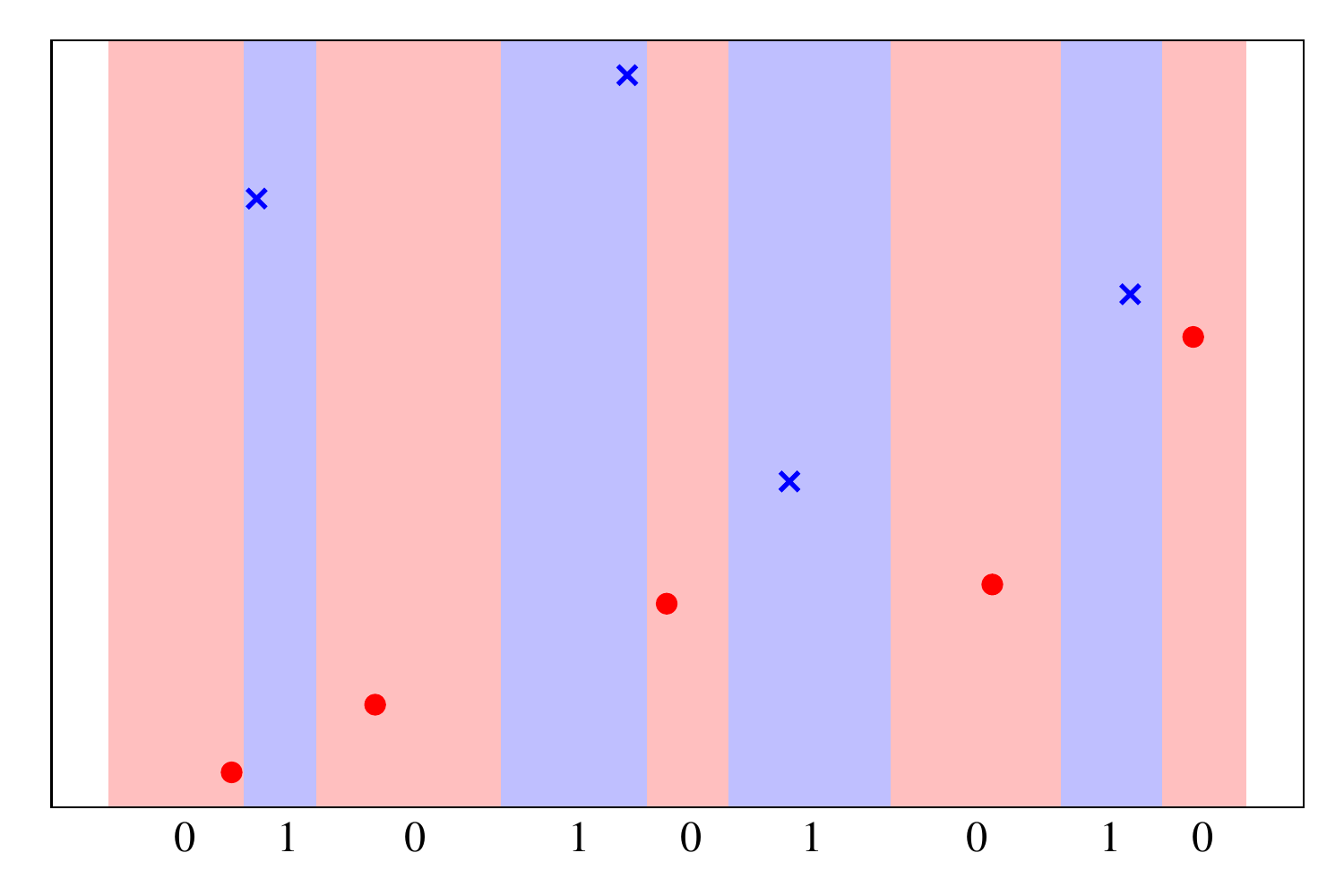}
    \caption{Example of overfitting inspired by \cite{Hanna2019-mi}. Only one encoder is needed. For points $\mathbf{x}$ in the blue region, $\mathcal{E}_1(\mathbf{x}) = y(\mathbf{x}) = 1$. Similarly, for those in the red regions, $\mathcal{E}_1(\mathbf{x}) = y(\mathbf{x}) = 0$. This scheme achieves zero classification error on the training set.}
    \label{fig:overfitting}
\end{figure}

In practice, only $\mathcal{T}$ is available, not $P$. Therefore, we minimize a regularized estimator of equation ($\ref{opt-problem}$) by partitioning $\mathcal{T}$ into training, validation, and test sets. Quantization is learned on the training set, while performance is evaluated on the test set. The validation set is used to tune the number of bins, as discussed in section \ref{sec-methods}. Note that it is always possible to achieve zero training error by setting $\mathcal{E}_1(\mathbf{x}) = y(\mathbf{x})$. This is shown in Figure \ref{fig:overfitting}. Clearly, this scheme overfits.

Throughout this work, we assume that the classifier available at the decoder performs binary linear classification. Formally, a weight vector $\mathbf{w} = (w_1, \dots, w_d) \in \mathbb{R}^d$ is available such that the class label is assigned as $\hat{y}(\mathbf{x}) = \text{sign}(\langle \mathbf{w},\mathbf{x} \rangle)$. We assume $\mathbf{w}$ is a 45 degree hyperplane (i.e. $-w_1=w_2=\dots=w_d=1$), as the training and test sets can always be re-scaled accordingly.

\section{Related Work}
Previous work has investigated distributed quantization schemes for both classification \cite{Hanna2019-mi} and traditional settings \cite{Wang2017-dz}. More recently, \cite{Hanna2019-mi} introduced the family of \emph{on-the-line} quantizers for linear binary classification in $\mathbb{R}^2$, which uses the same encoder for both dimensions. The core of the algorithm works by noting that there is a finite set of boundaries that affect the quantization error for a fixed training set of size $N$; which are the $2N$ coordinates of the data points. A greedy optimization strategy is given that finds an optimal solution to problem ($\ref{opt-problem}$), under the constraint that $\mathcal{E}_1 = \mathcal{E}_2$ and $P$ is the empirical distribution of the training set. It works by exhaustively searching the set of potential boundaries and keeping those which minimize the loss. To avoid overfitting, the authors impose continuity conditions on the quantization regions. The computational complexity is shown to be $\mathcal{O}(N^2 2^R)$, where $R=R_1=R_2$ is the rate.

This configuration can be viewed as related to distributed estimation and detection with communication-constrained links. See, \cite{chamberland2007wireless}, \cite{luo2005universal} and references therein. In \cite{longo1990quantization}, scalar quantization was studied. Many of these works assume that the sensor measurements are independently distributed given the detection hypothesis. Further, it is assumed that these conditional distributions are known. In our framework neither this conditional independence nor the knowledge of these distributions is assumed.

Several recent works focused on distributed probability estimation, property testing and simulation \cite{han2018geometric}, \cite{diakonikolas2017communication}, \cite{acharya2018distributed}. However, these works assume that each node observes all features while we focus on the case where each node observes a single feature.

Applying the pre-trained classifier over the quantized features can be viewed as functional compression problem. Distributed compression for functional computation with distortion has been studied (for example) in \cite{wagner2010distributed}, \cite{doshi2010functional}. Extending the ideas of \cite{Hanna2019-mi} our suggested algorithm performs quantization for unknown source distribution and without any apriori knowledge of the classifier function.

\section{Background}
\subsection{Distributed Lloyd-Max}\label{dist-lloyd-max}
Vanilla non-distributed Lloyd-Max \cite{lloyd1982least} is an iterative algorithm that computes optimal quantization boundaries and reconstruction points with respect to a loss function. It requires knowledge of the data distribution, but can be used with a training set by considering its empirical distribution. It alternates between two optimization steps. First, given an initial set of reconstruction points, it computes the bins by assigning each data point to the reconstruction point that minimizes the loss (bin step). Then, using the newly computed bins, it updates the reconstruction points to those which minimize the expected loss for each bin (reconstruction step). This method is known to converge to a local minimum of the loss. A good minima can be found by running the algorithm on random initialisations. Note that binning happens at the encoder, and reconstruction at the decoder.

In \cite{Rebollo-Monedero2005-uc}, a distributed version of Lloyd-Max is introduced. They provide conditions under which the loss function converges to a local minimum. The algorithm is similar to vanilla non-distributed Lloyd-Max, in the sense that it also alternates between two steps. However, to account for the distributed nature of the problem, it performs the bin step locally at each encoder while holding all other encoders fixed. The reconstruction step follows as in the non-distributed setting, since it happens at the decoder side.

\subsection{Generalization and Regularization}\label{subsec-mlbasics}
In our context, a learning algorithm $\mathcal{A}$ is a function that receives a training set of i.i.d\ points $\mathcal{T} = \{\mathbf{x}^{(i)}\}_{i=1}^N \sim P^N$ and outputs a distributed quantization scheme $\mathcal{A}(\mathcal{T}) = (\mathcal{E}, \mathcal{D})$, where $\mathcal{E}=(\mathcal{E}_1, \dots, \mathcal{E}_d)$ are the encoders and $\mathcal{D}$ the decoder. A loss function $\mathcal{L}$, such as the reconstruction or 0-1 loss, is used to quantify the performance. Given a fresh sample $\mathcal{T}^\prime \sim P^N$, we define two important quantities,

\begin{equation}
    \begin{aligned}
       \text{Training error} &\quad \mathcal{L}(\mathcal{A}(\mathcal{T}), \mathcal{T})\\
       \text{Test error}     &\quad \mathcal{L}(\mathcal{A}(\mathcal{T}), \mathcal{T^\prime}).
    \end{aligned}
\end{equation}

Note that in both cases $\mathcal{A}$ is trained on the training set $\mathcal{T}$, what differs is on which dataset the loss is evaluated. An algorithm is said to generalize well if both quantities are always similar in value.

When the training error is low and testing error is high, we say that the algorithm has overfitted. This can happen when $\mathcal{A}(\mathcal{T})$ has enough complexity that it outputs a quantization scheme which memorizes the training set (see Figure \ref{fig:overfitting}) \cite{shalev2014understanding}. Regularization can be seen as a way to reduce overfitting in $\mathcal{A}$ by augmenting the loss function such that it penalizes highly complex quantization schemes \cite{shalev2014understanding}.

In many practical cases, part of the training set $\mathcal{T}$ is held-out (i.e. not used during training) and is used to tune \emph{hyperparameters} \cite{shalev2014understanding}. This third dataset is called the validation set $\mathcal{T}_V$. Confusingly, the remaining dataset $\mathcal{T}_T = \mathcal{T}\setminus\mathcal{T}_V$ is also called the training set. Hyperparameters, in our context, can be understood as any parameter that is not optimized during training. Thus, the train/test paradigm is augmented to train/validate/test. Multiple quantizers, with different hyperparameter values, are trained on $\mathcal{T}_T$, and the one which performs best on $\mathcal{T}_V$ is chosen. Test loss is evaluated on $\mathcal{T}^\prime$, as before. This setting is employed in RCAQ, and is discussed in section \ref{sec-methods}.

\section{Methods}\label{sec-methods}

Our algorithm is a direct application of distributed Lloyd-max presented in \cite{Rebollo-Monedero2005-uc} and discussed in subsection \ref{dist-lloyd-max}. As the name entails, RCAQ works by regularizing the loss function with a term that penalizes bad reconstruction. The intuition stems from viewing this as a continuity condition in the data space. Points that are close together should tend to be quantized similarly. For this to be possible, we must add an extra step in the decoding process that maps the quantization indices $\mathcal{E}(\mathbf{x})$ to a reconstruction point $\mathcal{D}(\mathcal{E}(\mathbf{x})) = \hat{\mathbf{x}} \in \mathbb{R}^d$ before the classification function $\hat{y}$ is applied; as in traditional quantization.

\subsection{Algorithm}
\begin{figure}[h]
\centering
\includegraphics[width=\columnwidth]{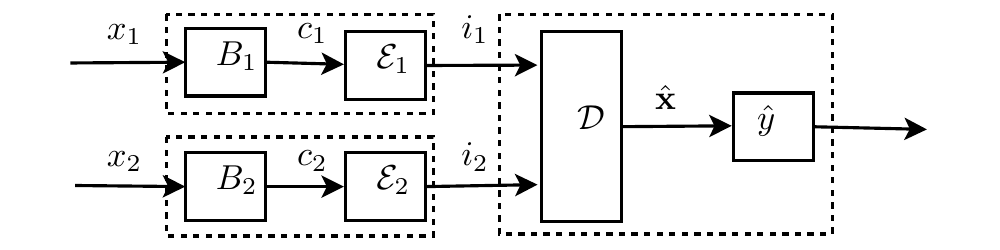}
\caption{RCAQ high level diagram. Data point $\mathbf{x} = (x_1, x_2)$ is mapped to bin centres $(c_1, c_2)$, and then to integer quantization indices $(i_1, i_2) \in [2^{R_1}]\times[2^{R_2}]$. The decoder deterministically selects a reconstruction point $\hat{\mathbf{x}}$, and applies the classifier $\hat{y}$.}
\label{fig:diagram}
\end{figure}

RCAQ is a two-stage quantizer. Encoders perform binning before quantization. Formally, each encoder has a binning function $B_i: \mathbb{R} \rightarrow \mathbb{R}$ that maps the data points to evenly spaced bins. The bin centres are then quantized with $\mathcal{E}_i$ before being sent to the decoder. The functions $\{B_i\}_{i=1}^d$ are \emph{not} optimized during training. Instead, an exhaustive search is performed over the number of bins during the validation phase discussed in sub-section \ref{subsec-mlbasics}. Thus, each encoder is represented by the pair $(B_i, \mathcal{E}_i)$, and the decoder by $(\mathcal{D}, \hat{y})$. A diagram is shown in Figure \ref{fig:diagram}.

Our loss function is
\begin{equation}\label{rcaq}
        \mathcal{L}_\mathcal{T}(\mathcal{E}, \mathcal{D}, \hat{y}) = \frac{1}{N}\sum_{\mathbf{x} \in \mathcal{T}} \left(\gamma\mathbbm{1}[\hat{y}(\hat{\mathbf{x}}) \neq y(\mathbf{x})] + (1-\gamma)\norm{\hat{\mathbf{x}} - \mathbf{x}}^2\right)
\end{equation}
where $\mathcal{T}$ is the training set, $N = \lvert \mathcal{T} \rvert$ the training set size, $0 \leq \gamma \leq 1$ controls the strength of regularization, $\norm{\hat{\mathbf{x}} - \mathbf{x}}^2$ is the reconstruction penalty, and $\mathbbm{1}$ the indicator function. For $\gamma=1$, $\mathcal{L}$ becomes the 0-1 loss, while for $\gamma=0$ it is the original Lloyd-max distortion term (i.e. MSE).

A small number of bins acts as a second regularizer, as it precludes the encoder from having highly discontinuous quantization regions. Larger values for $b$ increase the resolution at the encoder, making it possible to have discontinuous regions and therefore less distortion. However, in our experiments we found that the algorithm sometimes opts for discontinuity even for a relatively small $b \approx 16$. The pseudo-code for the optimization algorithm is presented and discussed next.
\begin{figure*}
    \centering
    \subfigure{\includegraphics[width=\columnwidth, trim=15em 0 12em 0, clip]{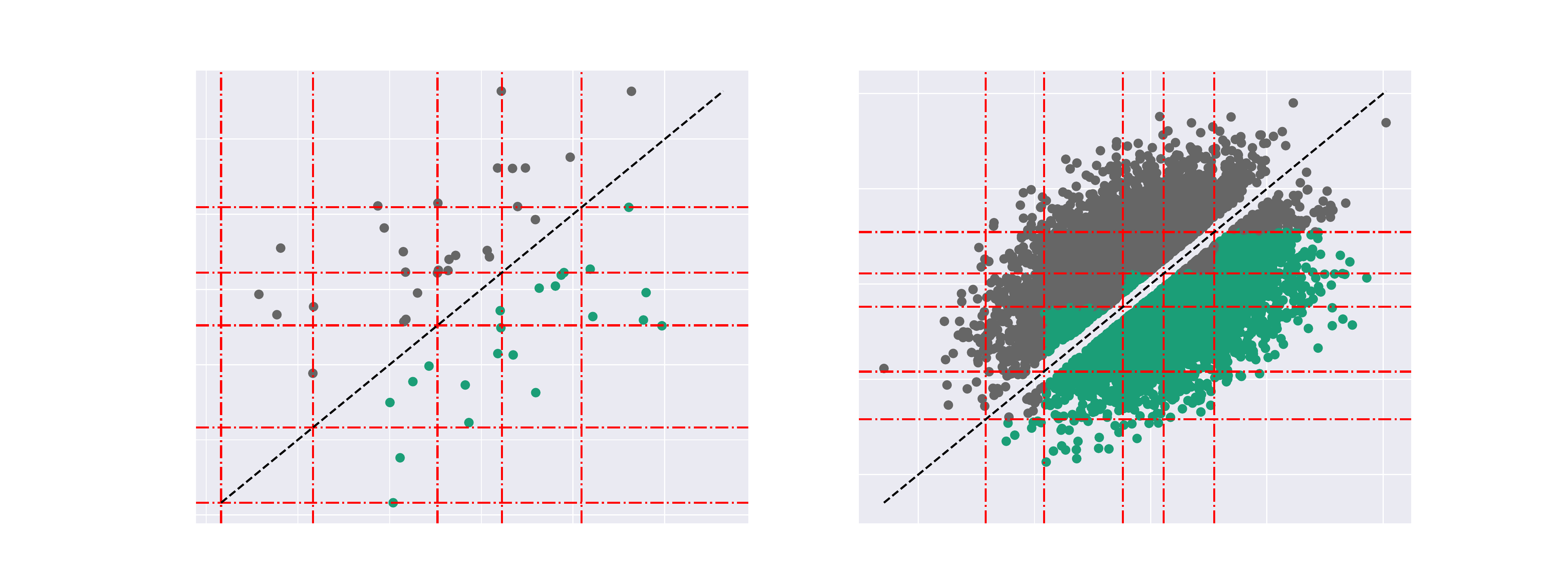}}\quad
    \subfigure{\includegraphics[width=\columnwidth, trim=15em 0 12em 0, clip]{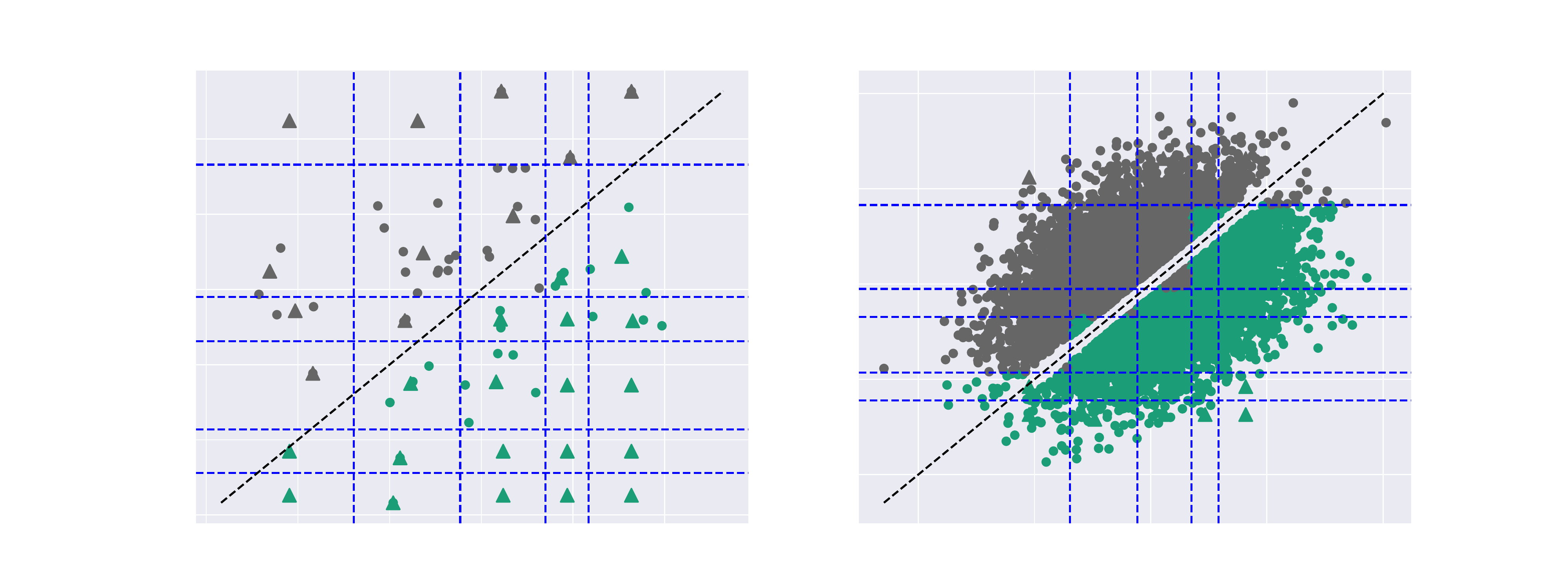}}
    \caption{Quantization bins for on-the-line (left, orange) and RCAQ (right, blue) for both training ($50$ points) and test sets ($10,000$ points). Class labels are represented by the color of each point (green or grey). Training loss for on-the-line and RCAQ are $0$ and $0.02$, respectively; while testing loss is $0.08$ and $0.05$. Reconstruction points are shown as triangles with their respective class labels corresponding to their colors. The number of bins for RCAQ is $10$, and the integer quantization index for each bin is shown in blue for each encoder. On-the-line does not use reconstruction points, as it assigns class labels to each quantization cell directly.}
    \label{fig:example}
\end{figure*}

\begin{algorithm}\label{alg-rcaq}
\SetAlgoLined
Given rate $R$ and training set $\mathcal{T}$\;
Partition $\mathcal{T}$ into $\mathcal{T}_T$ and validation set $\mathcal{T}_V$\;
Consider $R_i=R$ and $b_i=b$ for all $i \in \{1, \dots, d\}$\;
\For{number of bins $b \in \{1, 2, \dots, b_{\max}\}$}{
    Initialize encoders $\{(B_i, \mathcal{E}_i)\}_{i=1}^d$\;
    Initialize decoder $(\mathcal{D}, \hat{y})$\;
    \While{$\mathcal{L}_{\mathcal{T}_T}(\mathcal{E}, \mathcal{D}, \hat{y})$ has not converged}{
        \For{$i \in \{1, \dots, d\} $}{
            $\mathcal{E}_i \gets \arg\min\limits_{\mathcal{E}_i} \mathcal{L}_{\mathcal{T}_T}(\mathcal{E}, \mathcal{D}, \hat{y})$\;
         }
         $(\mathcal{D}, \hat{y}) \gets \arg\min\limits_{(\mathcal{D},\hat{y})} \mathcal{L}_{\mathcal{T}_T}(\mathcal{E}, \mathcal{D}, \hat{y})$\;
     }
    Evaluate 0-1 loss on validation set $\mathcal{T}_V$\;
    }
Output codec that achieved lowest 0-1 loss on $\mathcal{T}_V$\;

\caption{Reg. Classification-Aware Quantization}
\end{algorithm}

\subsection{Optimizing the encoder}
Algorithm \ref{alg-rcaq} assumes equal rates and number of bins in all dimensions, but this can be trivially extended. The inner optimization loop updates $\mathcal{E}_i$ by exhaustively searching which integer should be mapped to each bin. The overall computational complexity, if implemented naively, is $\mathcal{O}(Id\sum_{b=1}^{b_{\max}} 2^{Rb})$ where $R$ is the rate, $b$ the number of bins, and $I$ is the number of iterations to convergence. However, it is possible to parallelize over the bins by noting that the integer assigned to bin $i$ will not affect the loss incurred at bins $j \neq i$. In other words, the bins are independent during optimization. The outer validation loop can also be done in parallel. Also, we found that convergence is fast and does not depend significantly on other quantities, and hence $I$ is fixed in our experiments.
In practice, this reduces the complexity to
\begin{equation}
    \mathcal{O}\left(d2^R\right).
\end{equation}
When compared to the family of on-the-line quantizers ($d=2$), there is a reduction in complexity of $\mathcal{O}(N^2)$. The order in which the encoders are optimized matters, as the optimization step for encoder $\mathcal{E}_k$ uses the previous, and already optimized, encoders $\{\mathcal{E}_i\}_{i=1}^{k-1}$. Therefore, it is not possible to parallelize over the inner optimization loop, unless we relax the algorithm and optimize subsets of $\{\mathcal{E}_1, \dots, \mathcal{E}_d\}$ simultaneously. We do not investigate this latter scenario in this work.

\subsection{Optimizing the decoder}
Optimizing the decoder $(\mathcal{D}, \hat{y})$ consists of updating the reconstruction point function $\mathcal{D}$ as in regular Lloyd-max, but with the loss function defined in equation (\ref{rcaq}). The classification function $\hat{y}$ is simply the classifier itself. Note that, during minimization, the 0-1 loss component will only be affected if the reconstruction point is pushed to the other side of the decision boundary. This implies that the loss function (\ref{rcaq}) is discontinuous with respect to the reconstruction point. Surprisingly, this does \emph{not} add complexity to the minimization process, as there are only two possible updates for any reconstruction point. Either $\hat{\mathbf{x}}$ is updated to the average of the data-points assigned to it (i.e. as in vanilla Lloyd-max, which considers only MSE), or, if
\begin{equation}
  \sum_{\mathbf{x} \in \mathcal{T}_T(\hat{\mathbf{x}})}\gamma\mathbbm{1}[\hat{y}(\hat{\mathbf{x}}^\prime) \neq y(\mathbf{x})] > \sum_{\mathbf{x} \in \mathcal{T}_T(\hat{\mathbf{x}})}(1-\gamma)\norm{\hat{\mathbf{x}}^\prime - \mathbf{x}}^2,
\end{equation}
$\hat{\mathbf{x}}$ must be pushed to the other side of the boundary; where $\mathcal{T}_T(\hat{\mathbf{x}})$ is the set of training points that are currently assigned to reconstruction point $\hat{\mathbf{x}}$. Intuitively, this must be done in a way such that the increase in MSE is minimal. It is easy to show that the optimal pushing direction is exactly that of the normal vector defining the hyperplane, i.e. a straight line to the boundary. Therefore, the update procedure for a single reconstruction point $\hat{\mathbf{x}}$ is
\begin{equation}
    \argmin_{\hat{\mathbf{x}}^\prime \in \{\hat{\mathbf{x}}, \hat{\mathbf{x}} + \mathbf{\Delta}\}}
    \sum_{\mathbf{x} \in \mathcal{T}_T(\hat{\mathbf{x}})} \left(\gamma\mathbbm{1}[\hat{y}(\hat{\mathbf{x}}^\prime) \neq y(\mathbf{x})] + (1-\gamma)\norm{\hat{\mathbf{x}}^\prime - \mathbf{x}}^2\right),
\end{equation}
where $\mathbf{\Delta} \propto \mathbf{w}$ is the vector such that $\hat{\mathbf{x}} + \mathbf{\Delta}$ crosses the boundary by some small additive constant (we used $10^{-6}$).

\section{Experiments}
We asses the generalization performance of RCAQ and on-the-line quantizers with bivariate and mixture Gaussian data; as a function of both rate and correlation. Experiments were run for $1 \leq d \leq 5$ to confirm that the number of iterations to convergence $I$ does not scale significantly with $d$. However, experiments shown here are limited to $d=2$, as on-the-line is defined only in this setting. Artificial margins are created at the separation hyperplane (i.e. classifier) by adding a small perturbation in the direction of the normal vector (see Figure \ref{fig:example} for a visual). Intuitively, since there are less points concentrating around the decision boundary, larger margins should result in better generalization performance \cite{shalev2014understanding}. Indeed, this was observed experimentally for both RCAQ and on-the-line.

Empirically, we found that RCAQ is \emph{not} sensitive to $\gamma$, which serves mostly to equalize the scales of the 0-1 and reconstruction losses. In all experiments, $\gamma = 0.95$.

In the following sections, we detail the experimental setup. The experiments presented are a subset of the totality performed, and were chosen for being good representatives of the overall performance of both RCAQ and on-the-line.

\subsection{Bivariate Gaussians}\label{exp-bivariate}
Bivariate Gaussians with zero mean, equal variances and correlation coefficient $\rho$ were used, resulting in the following covariance matrix
\begin{equation}
    \mathbf{\Sigma} = \begin{pmatrix}
                      1 & \rho \\
                      \rho & 1
                      \end{pmatrix}.
\end{equation}

\begin{figure}
  \centering
  \subfigure{\includegraphics[width=\columnwidth]{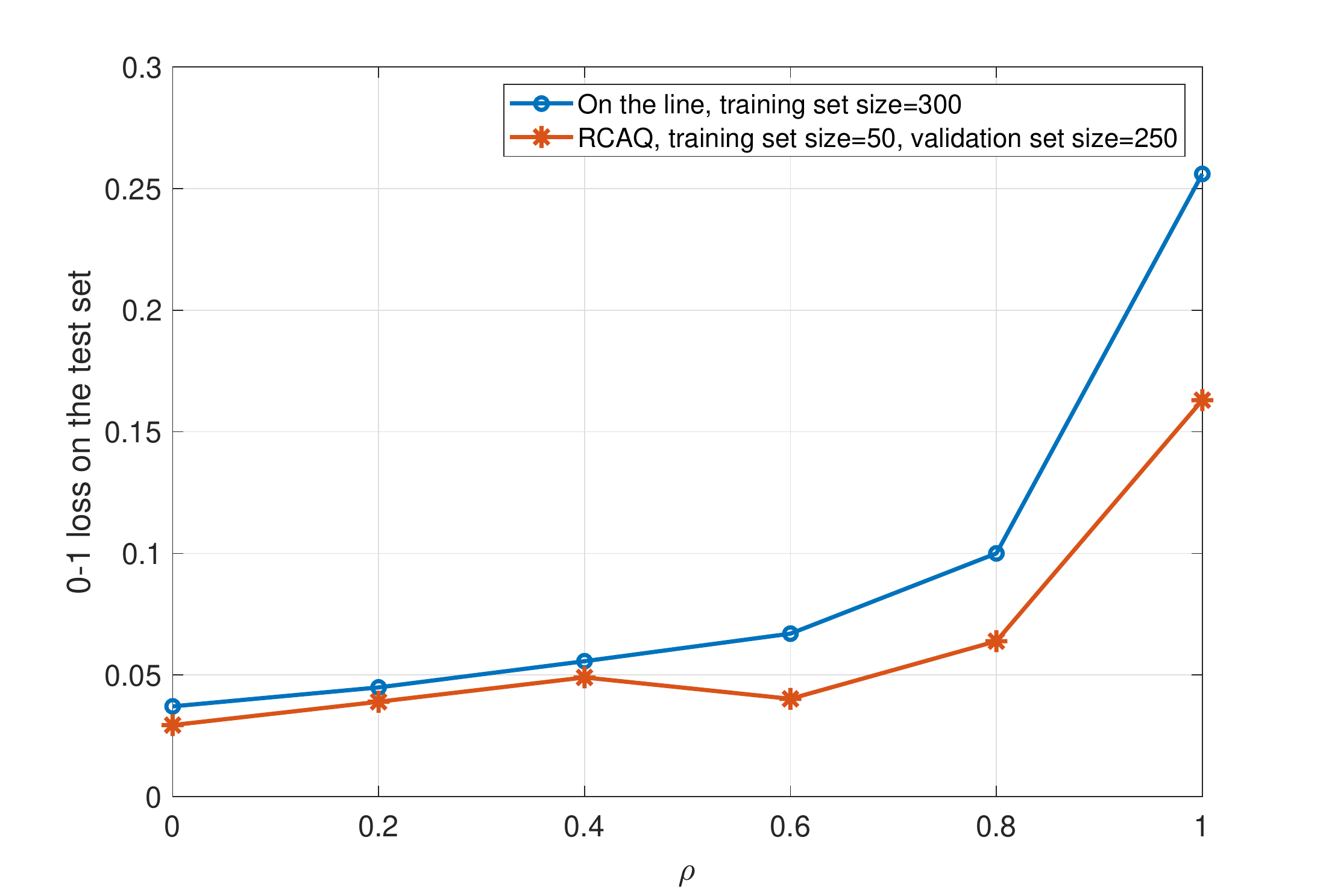}}
  \caption{0-1 loss (lower is better) for on-the-line (blue) and RCAQ (orange) quantizers. Test set of size $10,000$ was used. Results are shown as a function of the correlation coefficient for a bivariate Gaussian distribution.}
  \label{fig:bivariate-results}
\end{figure}

We generated $6$ training sets with correlation coefficient $\rho \in \{0, 0.2, 0.4, 0.6, 0.8, 1.0\}$ and training set size $N = 300$. Both on-the-line and RCAQ where trained on all $6$ and performance was measured on $6$ different test sets, each of size $10,000$, with the same values of $\rho$ as the training sets. The rate was fixed at $2^{R_1}=2^{R_2}=6$ integers per encoder (i.e. dimension) everywhere. The 0-1 loss for the test set is shown in Figure \ref{fig:bivariate-results} for both methods. For RCAQ, the train/validation split is set to $50/250$.

Our experiments indicate that RCAQ is, at the very least, competitive with on-the-line; but with a reduction in complexity of $\mathcal{O}(N^2)$. As $\rho$ increases, the loss increases in general, as more points concentrate along the decision plane making it harder to quantize. The effect of the margin is the opposite, increasing it pushes points away from the boundary, making quantization easier. This is true for both encoders, as previously mentioned.

In Figure \ref{fig:example}, we provide the quantization regions and indices for on-the-line and RCAQ for $\rho=0.4$ and $N=50$. Note that on-the-line does not use reconstruction points, as the quantization cells are assigned a class label directly. RCAQ uses reconstruction points, and chooses the class label by placing the reconstruction points below or above the hyperplane. The quantization cells for on-the-line that contain the hyperplane are always squares. To see this, note that the hyperplane is a 45 degree line and serves as the diagonal of the cell. RCAQ does not have this restriction, allowing those cells to be rectangles. This improves the generalization error, as the cells that intersect the hyperplane are the ones that can potentially misclassify points. In this example, on-the-line overfits as it achieves $0$ training and $0.08$ testing errors. RCAQ has a larger training error ($0.02$), but smaller testing error $(0.05)$.

\subsection{Gaussian mixtures}
We perform an experiment by fixing the rate of both RCAQ and on-the-line with data from a Gaussian mixture. The mean was uniformly sampled between $0$ and $1$, and a fixed diagonal covariance matrix $10^{-2}\mathbb{I}_d$ was used, where $\mathbb{I}_d$ is the identity matrix with $d$ rows and columns. The dataset sizes were $1,000$ for training and $10,000$ for testing. RCAQ's train/validation split was $300/700$.

Caution is needed when applying on-the-line at high rates. On-the-line iterates over the training data and inserts boundaries that minimize the training error. Therefore, if the training error reaches zero, it will stop adding boundaries even if more rate is available. To avoid this, and to guarantee a fair comparison, we evaluated on-the-line only for large dataset sizes ($N=1,000$). Still, RCAQ clearly outperforms on-the-line on the test loss.
\begin{figure}
  \centering
  \subfigure{\includegraphics[width=\columnwidth]{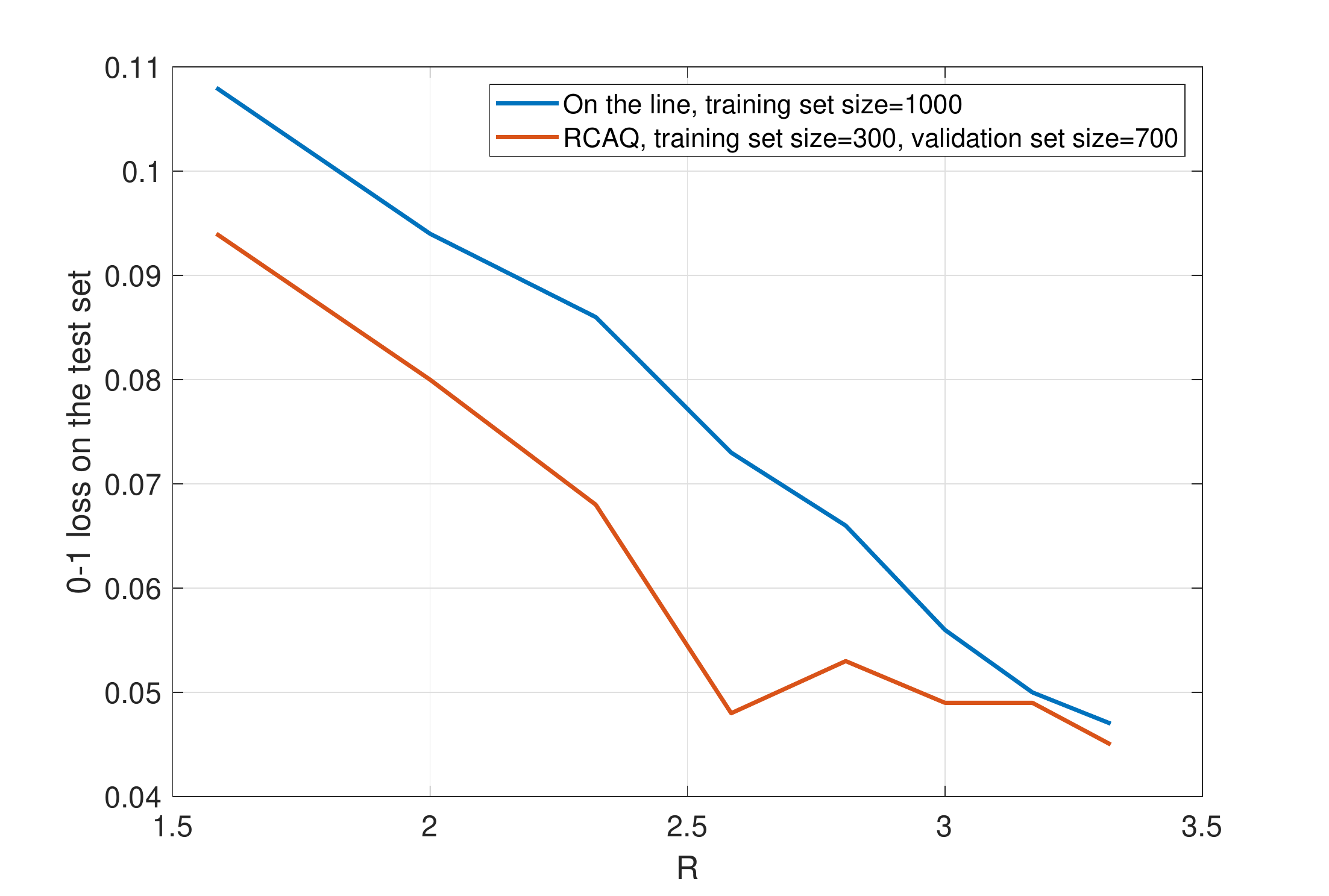}}
  \caption{0-1 loss (lower is better) for on-the-line (blue) and RCAQ (orange) quantizers. Test set of size $10,000$ was used. Results are shown as a function of the rates $R_1 = R_2 = R$ (i.e. $2^R$ bins at each encoder) for a Gaussian mixture distribution.}
\end{figure}

\subsection{Gauging the potential of RCAQ}

\begin{figure}
  \centering
  \subfigure{\includegraphics[width=\columnwidth]{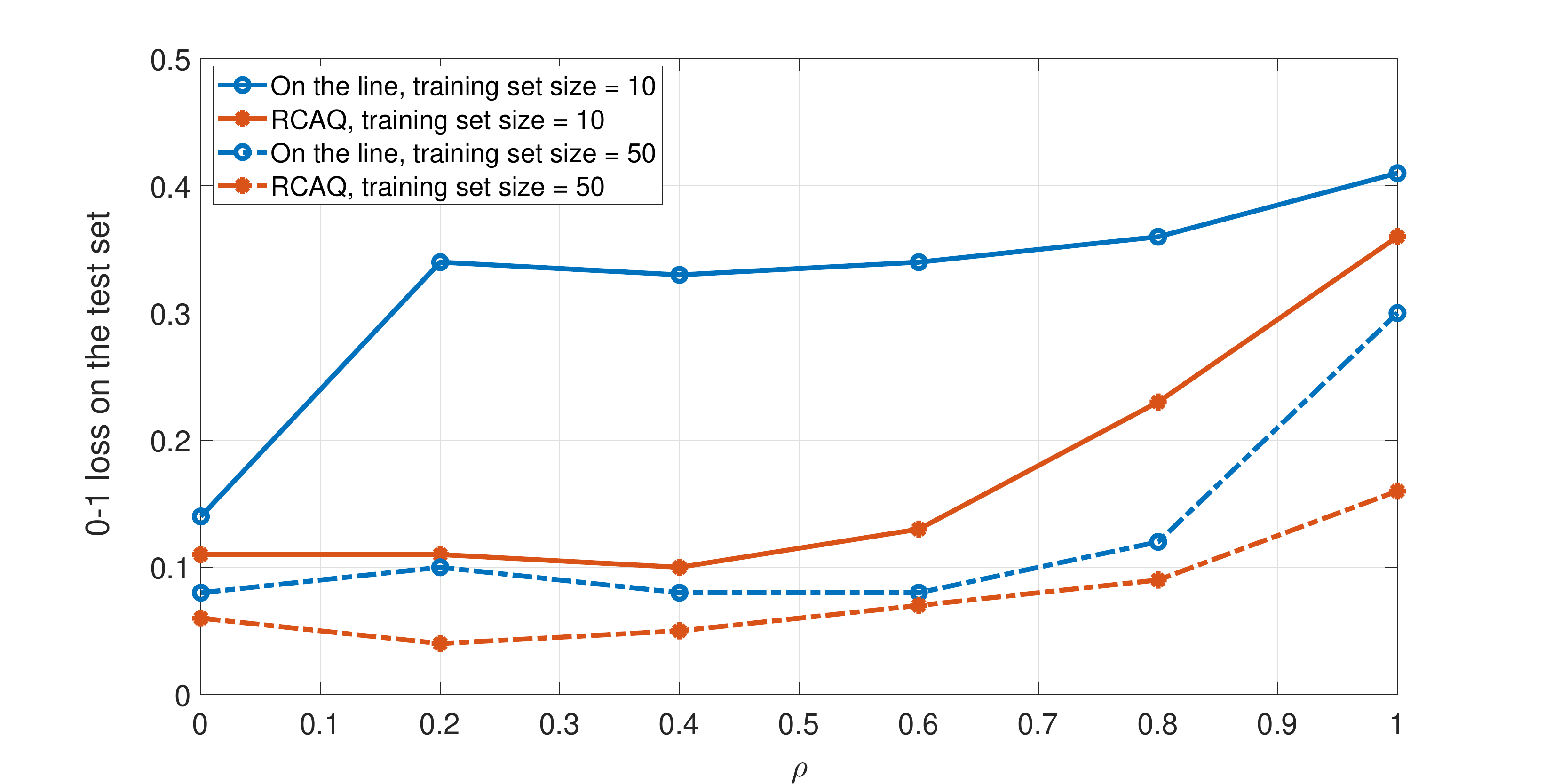}}
  \caption{Testing loss for on-the-line (blue) and RCAQ (orange) quantizers. Two training sets of sizes $10$ (solid) and $50$ (dashed) where used. Results are shown as a function of the correlation coefficient for a Bivariate Gaussian distribution. Both methods achieved training loss close to zero, but RCAQ was slightly higher.}
  \label{fig:results}
\end{figure}

As shown in previous sections, RCAQ performs competitively with on-the-line, with significantly less complexity. However, we postulate that RCAQ can perform better for small training set size.

A major component of RCAQ is the validation step which is used to find the number of evenly spaced bins for each encoder; $\{b_i\}_{i=1}^d$. To motivate future work, we performed experiments to understand if there exists a good set $\{b_i^*\}_{i=1}^d$ to begin with, i.e. to asses the potential of RCAQ with respect to the number of evenly spaced bins. To do this, we forgo the validation step, fix $b_1 = b_2 = \dots = b_d = b$, and exhaustively search over $1 \leq b \leq 32$ by \emph{evaluating on the test set}. Clearly, this setup is not feasible in practice, as we are optimizing over the test set. However, it does allow us to gauge the potential performance of RCAQ. On-the-line was trained only with the training set, hence the results shown are representative of its performance. Results are shown in Figure \ref{fig:results}. Our experiments indicate that RCAQ can potentially outperform on-the-line. With few training points, on-the-line is not able to generalize, as the algorithm works by inserting quantization boundaries on the coordinates of the training points themselves. This is reflected in the gap between the losses of RCAQ and on-the-line, as the potential advantage incurred by RCAQ diminishes as the training set size increases. Also, RCAQ does not limit the solution to $\mathcal{E}_1 = \mathcal{E}_2$, which allows more flexibility.

\section{Conclusion}
In this work we proposed a new quantization algorithm for distributed classification called \emph{Regularized Classification-Aware Quantization} (RCAQ). We provide experiments that show that RCAQ can generalize to out-of-sample data from few training points. Through vectorization, an implementation is provided with computational complexity $\mathcal{O}(d 2^R)$, where $d$ is the number of encoders and $R$ is the rate at each encoder. RCAQ is competitive with current baselines such as on-the-line, but with significantly lower computational complexity, as it does not loop over the dataset to pick boundaries.

We provide experiments that motivate future work, by showing that there exist good configurations of evenly-spaced boundaries that can outperform existing methods. Finding them, however, is left as an open problem. More experiments are needed to asses if this scheme will scale to high dimensional datasets.

\bibliography{bibliography}{}
\bibliographystyle{plain}

\end{document}